# A Structured Sparse Neural Network and Its Matrix Calculations Algorithm

Seyyed Mostafa Mousavi Janbeh Sarayi [1*], Mansour Nikkhah Bahrami [2]

[1*]Department of Mechanical and Aerospace Engineering, State University of New York at Buffalo, Buffalo, New York
[2]Department of Mechanical Engineering, College of Engineering, University of Tehran, Tehran, Iran.
* Corresponding author: E-mail: smousavi@buffalo.edu

**Abstract**

Gradient descent optimizations and backpropagation are the most common methods for training neural networks, but they are computationally expensive for real time applications, need high memory resources, and are difficult to converge for many networks and datasets. [Pseudo]inverse models for training neural network have emerged as powerful tools to overcome these issues. In order to effectively implement these methods, structured pruning may be applied to produce sparse neural networks. Although sparse neural networks are efficient in memory usage, most of their algorithms use the same fully loaded matrix calculation methods which are not efficient for sparse matrices. Tridiagonal matrices are one of the frequently used candidates for structuring neural networks, but they are not flexible enough to handle underfitting and overfitting problems as well as generalization properties. In this paper, we introduce a nonsymmetric, tridiagonal matrix with offdiagonal sparse entries and offset sub and super-diagonals as well algorithms for its [pseudo]inverse and determinant calculations. Traditional algorithms for matrix calculations, specifically inversion and determinant, of these forms are not efficient for large matrices. A decomposition for lower triangular matrices is developed and the original matrix is factorized into a set of matrices where their inverse matrices are calculated. For the cases where the matrix inverse does not exist, a least square type pseudoinverse is provided. The present method is a direct routine, i.e., executes in a predictable number of operations which is tested for randomly generated matrices with varying size. The results show significant improvement in computational costs specially when the size of matrix increases.

# 1. Introduction

Gradient Descent with backpropagation is the traditional approach for training neural networks. However, they are slow and difficult to converge because of their iterative nature, there are several hyperparameters to tune, gradients can vanish or explode, and they need multiple visits (epochs) of the dataset. Inverse or pseudoinverse learning processes have advantages over variation of Gradient Descent algorithms [1,2] like non-iterative based search, one epoch of dataset, and better convergence. Despite their success, they could be computationally expensive specially when it comes to heavy networks.

Structured pruning and subsequent sparse matrix calculations are effective for optimizing different machine learning algorithms [3–5] and specially heavy neural networks [6–9]. In neural network applications, structured pruning result in sparse, structured weight matrices to calculate neurons through its system of equations. Tridiagonal weight matrices offer a structure which could be implemented for these purposes, but they are limited in the way they connect neurons. By adding more degrees of freedom to these matrices through off diagonal entries and offset diagonals, it is possible to extended their capabilities to more accurate and at the same time efficient models. Not only these matrices are applicable to neural network pruning, but also they appear in many applied mathematics problems like digital image processing [10], time series analysis [11], discrete Fourier transform (DFT) [12], spline approximation [13], cyclic codes for error correction [14], computational dynamics [15], and finite difference solution of one-dimensional elliptic equations subjected to periodic boundary condition [16]. One drawback of adding more degrees of freedom to a network and having these matrices are that matrix calculations for the produced matrices are not efficient and most of the present algorithms are designed either for fully loaded matrices or specific ones like tridiagonal matrices.

There are studies concerning more efficient algorithms for sparse, structured matrices and system of equations solutions. Solving linear systems with tridiagonal and diagonal dominant symmetric Toeplitz matrices was studied by Rojo [17]. In this research, the idea is to decompose a Toeplitz matrix into a sum of a pair of matrices so that one of them has a simple LU-type Toeplitz like decomposition, which can be computed with constant cost and allows us to solve the tridiagonal system in $O(n)$. Vidal and Alonso in another study [18] revised Rojo's algorithm for symmetric Toeplitz tridiagonal equations. Broughton and Leader [19] represented an analytical formula for the inverse of a symmetric circulant tridiagonal matrix as a product of a circulant matrix and its transpose. Nemani and Garey [20] presented a new stable algorithm for the solution of tridiagonal circulant linear systems of equations. Recursive algorithms for finding the inverse of tridiagonal, anti-tridiagonal, pentadiagonal, and anti-pentadiagonal matrices were presented in studies by El-Mikkawy and Rahmo [21] [22]. Furthermore, nonsymmetric tridiagonal Toeplitz systems have been studied by Garey and Shaw [23]. By making transformation matrices, Zheng and Shon [24] studied the determinant and inverse of a generalized Lucas skew circulant matrix. The determinant and inverse of the Tribonacci skew circulant matrix were presented based on making the transformation matrices in a study by Jiang and Hong [25]. Moreover, there are a number of studies in literature that studied blocked Toeplitz and circulant systems. In some cases where multi-dimensional problems are concerned the matrices of the resulting linear systems are block Toeplitz and circulant matrices [26,27]. However, these block Toeplitz and circulant matrices are transformed into block diagonal matrices with Toeplitz and circulant blocks on the diagonal leading to the solution of Toeplitz and circulant linear systems of lower orders. As such, Fuyong [28] applied a direct method to solve the inverse problem for a block circulant banded system of linear equations by using the kernel solution of linear equations.

Previous studies have mostly worked on specific types of structures which limits the degrees of freedom needed for neural networks learning process and generalization problem. This study aims to introduce an novel approach for constructing structured, sparse neural networks based on a sparse matrix and introducing an efficient algorithm to find inverse and pseudoinverse of the matrix which is a more flexible structured matrices with extended degrees of freedom: nonsymmetric tridiagonal matrices with off diagonal entries and offset diagonals. To this purpose, the introduced matrix is factorized and their closed-form inverses are found using another decomposition of lower triangular matrices. This decomposition for lower triangular matrices, introduced for the first time, is derived from a general decomposition method [29]. The factor matrices produce a LLU decomposition that can be compared with LU decomposition algorithms, namely Crout's algorithm which takes $O(n^3)$ operations to invert a matrix, the same count as Gauss-Jordan matrix inversion [30]. In these methods, one needs to solve $n$ linear systems with right-hand side the identity matrix, namely $n$ unit vectors which are the columns of the identity matrix. Although there are some solution procedures specific for tridiagonal matrix inversion which are faster, the matrix of present work has offdiagonal elements that make these special algorithms fail. On the other hand, using formulations provided in this research, factor matrices are found in $O(n)$ and the inverse of the factor matrices in $O(n^2)$ arithmetic operations. However, the order of operations needed for finding the inverse of the matrix $A$ is dominated by the order of operations needed for multiplying factor matrices. While the product of two $n \times n$ matrices over a field can naturally be computed in $O(n^3)$ arithmetic operations, more efficient algorithms exist for large $n$. Strassen [31] showed in 1969 that $O(n^{2.81})$ arithmetic operations are enough. Coppersmith and Winograd [32] provided an algorithm that can multiply two $n \times n$ matrices in $O(n^{2.3754770})$ time. Stothers [33] gave an improvement to the algorithm, $O(n^{2.374})$. In 2011, Williams [34] combined a mathematical short-cut from Stothers' research with her own insights and automated optimization on computers, improving the bound to $O(n^{2.3728642})$. Le Gall [35] simplified the methods of Williams and obtained an improved bound of $O(n^{2.3728639})$. The determinant of matrix $A$ is also found that could be used for checking convergence. Performance of the algorithm is tested by comparing the computational cost in finding the inverse of random matrices against the most efficient algorithms and the results showed considerable improvements.

## 2. Structured sparse neural networks

Considering a single layer feedforward network with M hidden nodes and activation function Z, mathematical model for $N$ distinct samples and target ($X \in R^{N \times n}, T \in R^{N \times m}$) yields:

$$\sum_{i=1}^{M} \alpha_i Z(w_i . x_j + b_i) = t_i, \quad j = 1, \dots, N \tag{1}$$

assuming there exists $W \in R^{M \times n} (w_i = [w_{i1}, \dots, w_{in}]^T)$ input-hidden layer weights and $A \in R^{M \times m} (\alpha_i = [\alpha_{i1}, \dots, \alpha_{in}]^T)$ hidden layer-output weights, $B$ bias for hidden nodes, and linear output nodes where $\sum_{i=1}^{M} ||\alpha_i Z(w_i . x_j + b_i) - t_i|| = 0$. The above equation in matrix form could be written as:

$$G(w_1, , \dots w_M, b_1, \dots, b_M) A = T \tag{2}$$

Therefore, the training procedure reduces to an optimization problem of the least square form as:

$$\min (||G(w_1, , \dots w_M, b_1, \dots, b_M) A - T||) \tag{3}$$

which a closed form solution could be achieved ($G^{-1}$ is the inverse of the matrix if it exists or pseudoinverse if it does not):

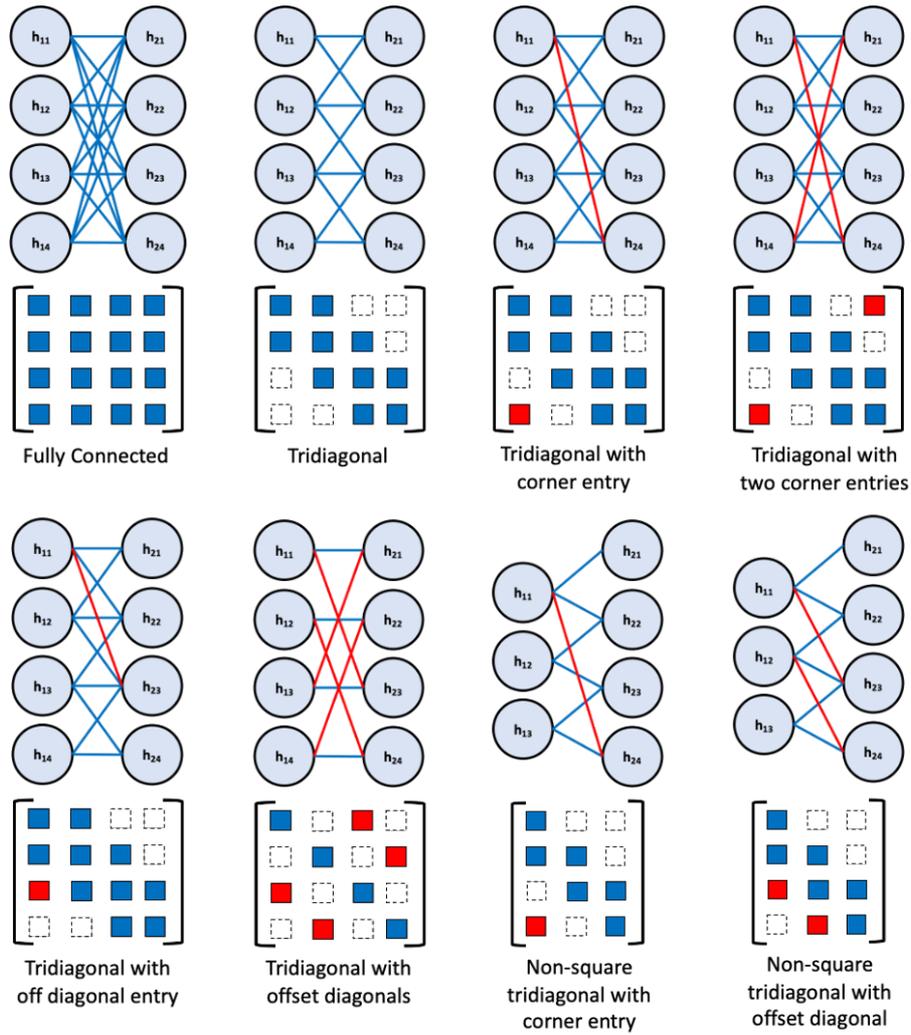

Figure 1. Proposed sparse structuring of neural networks with their extended degrees of freedom

$$A = G^{-1}T \tag{3}$$

However, since in most cases number of hidden nodes and samples are different and finding an inverse or a pseudoinverse is computationally expensive for fully connected networks (figure (1)-top left), we reduce the matrix calculations by structured pruning as shown in figure (1). Different strategies could be adopted for structured pruning, We used a tridiagonal matrix form as the basis pruning structure, and to add flexibility against underfitting and overfitting problems, we need to added more degrees of freedom to these matrices by offdiagonal entries and offset diagonals (red lines and red matrix entries in figure (1).

The matrix $A_{n \times n} = [a_{i,j}; i,j = 1,2,....,n]$ considered in this research is an almost tridiagonal nonsingular matrix of the form

$$a_{i,j} = \begin{cases} a_i & i = 1,2,\cdots,n \text{ and } j = i \\ b_j & i = 2,3,...n \text{ and } j = i-1 \\ c_i & i = 1,2,...n-1 \text{ and } j = i+1 \\ d_2 & i = 1 \text{ and } j = n \\ d_1 & i = n \text{ and } j = 1 \\ 0 & otherwise \end{cases} \quad (4)$$

In this matrix, considering $a_i = a$, $d_2 = b_i = b$ and $d_1 = c_i = c$ converts the matrix into a nonsymmetric tridiagonal circulant matrix. Considering $a_i = a$, $b_i = b$, $c_i = c$ and $d_1 = d_2 = 0$ ($i = 1 \cdots n-1$) converts the matrix to a nonsymmetric tridiagonal Toeplitz matrix. Finally, $d_1 = d_2 = 0$ converts the matrix to a nonsymmetric tridiagonal matrix. A circulant matrix is a special type of Toeplitz matrix. A $n \times n$ matrix $A_{n \times n} = [a_{i,j}; i,j = 1,2,....,n]$ is said to be Toeplitz if $a_{i,j} = a_{i+1,j+1}$. Toeplitz matrix is circulant if the matrix is row-wise wrap-around, or simply the subscripts are taken modulo $n$.

## 3. A decomposition method for lower triangular matrices

Any matrix $A_{n \times n}$ could be decomposed into $n$ matrices as $A = A_1 A_2 .... A_m ... A_n$ where $A_m = [a^m_{i,j}; i,j = 1,2,....,n]$ is:

$$a^m_{i,j} = \begin{cases} a_{i,j} & i = m \text{ and } j = 1,2,\cdots,m \\ \tau_{i,j} & i = 1,2,\cdots,m-1 \text{ and } j = m \\ 1 & i = j \text{ and } j = 1,\cdots n \text{ except for } i = j = m \\ 0 & otherwise \end{cases} \quad (5)$$

which has a form as:

$$A_m = \begin{bmatrix} 1 & 0 & \cdots & 0 & \tau_{1,m} & 0 & \cdots & 0 \\ 0 & 1 & \cdots & 0 & \tau_{2,m} & 0 & \cdots & 0 \\ \vdots & \vdots & \ddots & 0 & \vdots & \vdots & \vdots & \vdots \\ 0 & \cdots & 0 & 1 & \tau_{m-1,m} & 0 & \vdots & \vdots \\ a_{m,1} & a_{m,2} & \cdots & a_{m-1,m} & a_{m,m} & 0 & \vdots & \vdots \\ 0 & 0 & \cdots & 0 & 0 & 1 & \ddots & \vdots \\ \vdots & \vdots & \vdots & \vdots & \vdots & \vdots & \ddots & 0 \\ 0 & 0 & \cdots & \cdots & \cdots & 0 & 0 & 1 \end{bmatrix} \quad (6)$$

and the closed formula for the $\tau$ is:

$$\begin{cases} \tau_{i,k+1} = \dfrac{a_{i,k+1} - \begin{bmatrix} a_{i,1} \ldots a_{i,i-1} \end{bmatrix} \begin{bmatrix} \tau_{1,k+1} \ldots \tau_{i-1,k-1} \end{bmatrix}}{a_{i,i} - \begin{bmatrix} a_{i,1} \ldots a_{i,i-1} \end{bmatrix} \begin{bmatrix} \tau_{1,i} \ldots \tau_{i-1,i} \end{bmatrix}} \\ \tau_{s,k+1} = \tau_{s,k+1} - \tau_{s,i}\, \tau_{i,k+1} \end{cases} \quad (7)$$

for $i = 1,2,\ldots,n-1$, $s = 1,2,\ldots,i-1$ and $k = i, i+1, \cdots, N-1$. There is an interesting fact about this factorization; if we apply Gaussian elimination to $A^T$, we can obtain a lower triangular matrix $L$ and an upper triangular matrix $U$ as follows:

$$LA^T = U \quad (8)$$

or by transposing:

$$AL^T = U^T$$

From the triangular structure of $L^T$ and $U^T$ and the definition of the elements $\tau_{1,m}, \tau_{2,m}, \ldots, \tau_{m-1,m}$, it is found that:

$$L^T = \begin{bmatrix} 1 & -\tau_{1,2} & -\tau_{1,3} & \cdots & -\tau_{1,n} \\ 0 & 1 & -\tau_{2,3} & \cdots & -\tau_{2,n} \\ 0 & 0 & 1 & \ddots & \vdots \\ \vdots & \vdots & \ddots & \ddots & -\tau_{n-1,n} \\ 0 & 0 & \cdots & 0 & 1 \end{bmatrix} \quad (9)$$

What makes the application of this decomposition even simpler and more practical is that for lower triangular matrices all $\tau$ are zero and elements of the matrix $A_{n \times n}$ are the only elements of the factor matrices. Consider a lower triangular matrix of order $n$:

$$A_{n \times n} = \begin{bmatrix} a_{1,1} & 0 & 0 & \cdots & 0 \\ a_{2,1} & a_{2,2} & 0 & \cdots & 0 \\ a_{3,1} & a_{3,2} & a_{3,3} & \ddots & \vdots \\ \vdots & \vdots & \vdots & \ddots & 0 \\ a_{n,1} & a_{n,2} & a_{n,3} & \cdots & a_{n,n} \end{bmatrix} \qquad (10)$$

Applying the decomposition to this matrix yields

$$A_{n \times n} = \begin{bmatrix} a_{1,1} & 0 & \cdots & & 0 \\ 0 & 1 & \cdots & & \vdots \\ \vdots & \ddots & \ddots & \ddots & \vdots \\ \vdots & & \ddots & 1 & 0 \\ 0 & \cdots & \cdots & 0 & 1 \end{bmatrix} \begin{bmatrix} 1 & 0 & \cdots & \cdots & 0 \\ a_{2,1} & a_{2,2} & \ddots & \cdots & \vdots \\ 0 & 0 & 1 & \ddots & \vdots \\ \vdots & & \ddots & \ddots & 0 \\ 0 & \cdots & \cdots & 0 & 1 \end{bmatrix} \cdots \begin{bmatrix} 1 & 0 & \cdots & & 0 \\ 0 & 1 & \ddots & & \vdots \\ \vdots & \ddots & \ddots & \ddots & \vdots \\ 0 & \cdots & 0 & 1 & 0 \\ a_{n,1} & a_{n,2} & \cdots & \cdots & a_{n,n} \end{bmatrix} \qquad (11)$$

Simply, the inverse of matrix $A$ can be calculated as follows

$$A^{-1}{}_{n \times n} = \begin{bmatrix} 1 & 0 & \cdots & 0 & 0 \\ 0 & 1 & \ddots & \vdots & \vdots \\ \vdots & \vdots & \ddots & 0 & 0 \\ 0 & \cdots & 0 & 1 & 0 \\ -\dfrac{a_{n,1}}{a_{n,n}} & -\dfrac{a_{n,2}}{a_{n,n}} & \cdots & \cdots & \dfrac{1}{a_{n,n}} \end{bmatrix} \begin{bmatrix} 1 & 0 & 0 & \cdots & 0 \\ -\dfrac{a_{2,1}}{a_{2,2}} & \dfrac{1}{a_{2,2}} & 0 & \cdots & 0 \\ 0 & 0 & 1 & \ddots & \vdots \\ \vdots & \vdots & & \ddots & 0 \\ 0 & 0 & \cdots & 0 & 1 \end{bmatrix} \cdots \begin{bmatrix} \dfrac{1}{a_{1,1}} & 0 & \cdots & \cdots & 0 \\ 0 & 1 & \ddots & \ddots & \vdots \\ \vdots & \ddots & \ddots & \ddots & \vdots \\ \vdots & & \ddots & \ddots & 0 \\ 0 & \cdots & \cdots & 0 & 1 \end{bmatrix} \qquad (12)$$

A closed-form formula for the product of the Eq. (12) can be found but it is too lengthy and out he scopes of this study. In the following section the matrix of the present study is factorized into three lower triangular matrices. As the component matrices are sparse, the closed form formula for inverse of each of them could be found using this decomposition.

## 4. Formulation

The matrix we consider here is a nonsymmetric almost tridiagonal matrix introduced in Eq. (1) and is of the form

$$A_{n \times n} = \begin{bmatrix} a_1 & c_1 & 0 & \cdots & 0 & d_2 \\ b_1 & a_2 & c_2 & \ddots & \vdots & 0 \\ 0 & b_2 & \ddots & \ddots & 0 & \vdots \\ \vdots & 0 & \ddots & \ddots & c_{n-2} & 0 \\ 0 & \vdots & \ddots & b_{n-2} & a_{n-1} & c_{n-1} \\ d_1 & 0 & \cdots & 0 & b_{n-1} & a_n \end{bmatrix}_{n \times n}, \quad (|a_i| > |c_i| + |b_i|, \; b_i \neq 0) \quad (13)$$

where $|a_i| > |c_i| + |b_i|$, denoting the matrix is diagonal dominant. This matrix is factorized as

$$A = \Theta \Psi (\Re)^T \tag{14}$$

where $\Theta$, $\Psi$ and $\Re$ are three lower triangular sparse matrices and their exact analytical inverses are found.

The bidiagonal matrix $\Theta_{n \times n} = \left[ \theta_{i,j} ; i, j = 1, 2, \ldots, n \right]$ is of the form

$$\theta_{i,j} = \begin{cases} \dfrac{1}{\lambda_j} & i = j \text{ and } j = 1, 2, \cdots n \\ -\dfrac{1}{\lambda_i} & i = j+1 \text{ and } j = 1, 2, \cdots n-1 \\ 0 & \text{otherwise} \end{cases} \tag{15}$$

where all the entries are zero except for the elements of the main diagonal and subdiagonal as:

$$\Theta_{n \times n} = \begin{bmatrix} \dfrac{1}{\lambda_1} & 0 & \cdots & \cdots & 0 \\ -\dfrac{1}{\lambda_2} & \dfrac{1}{\lambda_2} & \ddots & \ddots & \vdots \\ 0 & -\dfrac{1}{\lambda_3} & \ddots & \ddots & \vdots \\ \vdots & \ddots & \ddots & \dfrac{1}{\lambda_{n-1}} & 0 \\ 0 & \cdots & 0 & -\dfrac{1}{\lambda_n} & \dfrac{1}{\lambda_n} \end{bmatrix}_{n \times n}$$

where $\lambda_j$ are obtained from the following relation:

$$\lambda_{j+1} = -\frac{(a_j \lambda_j + c_{j-1} \lambda_{j-1})}{b_j}, \quad (a_j \lambda_j + c_{j-1} \lambda_{j-1} \neq 0) \tag{16}$$

$$\lambda_0 = 0, \lambda_1 = 1, j = 1, 2, 3, \ldots, n-1$$

And the determinant $\det(\Theta)$ of this matrix is:

$$\det(\Theta) = \prod_{i=1}^{n} \frac{1}{\lambda_i} \tag{17}$$

The inverse $\Theta^{-1}_{n \times n} = [\sigma_{i,j}; i, j = 1, 2, \ldots, n]$ of the matrix $\Theta_{n \times n}$ is obtained using the decomposition method presented in section 2 as:

$$\sigma_{i,j} = \begin{cases} \lambda_j & i \geq j \\ 0 & j > i \end{cases} \tag{18}$$

that has the following form:

$$\Theta^{-1}_{n \times n} = \begin{bmatrix} \lambda_1 & 0 & \cdots & \cdots & 0 \\ \lambda_1 & \lambda_2 & \ddots & \ddots & \vdots \\ \vdots & \lambda_2 & \ddots & \ddots & \vdots \\ \vdots & \vdots & \ddots & \lambda_{n-1} & 0 \\ \lambda_1 & \lambda_2 & \cdots & \lambda_{n-1} & \lambda_n \end{bmatrix}_{n \times n}$$

It can be seen in Eq. (15) and Eq. (18) that the matrix $\Theta$ and its inverse have elements that can be calculated directly.

Now let matrix $\Psi_{n \times n} = [\xi_{i,j}; i, j = 1, 2, \ldots, n]$ be of the form

$$\xi_{i,j} = \begin{cases} 1 & i = j \text{ and } j = 1, 2, \cdots n \\ -\zeta_j & i = n \text{ and } j = 1, 2, \cdots n-1 \\ 0 & \text{otherwise} \end{cases} \tag{1}$$

where $\Psi_{n \times n}$ is a sparse matrix of the shape

$$\Psi_{n\times n} = \begin{bmatrix} 1 & 0 & \cdots & 0 & 0 \\ 0 & 1 & \ddots & \vdots & \vdots \\ \vdots & \ddots & \ddots & 0 & \vdots \\ 0 & \cdots & 0 & 1 & 0 \\ -\zeta_1 & -\zeta_2 & \cdots & -\zeta_{n-1} & 1 \end{bmatrix}_{n\times n} \quad (2)$$

In Eq. (19) $\zeta_j$ can be calculated as:

$$\zeta_j = -\frac{c_{j-1}\lambda_{j-1}\zeta_{j-1}}{a_j\lambda_j + c_{j-1}\lambda_{j-1}},$$
$$\zeta_1 = -\frac{d_1\lambda_n}{a_1}, \quad j = 2,3,\cdots n-1 \quad (3)$$

It is evident that the determinant $\det(\Psi)$ of this matrix is one.

The inverse $\Psi^{-1}_{n\times n} = \left[ \upsilon_{i,j}; i,j = 1,2,....,n \right]$ of $\Psi_{n\times n}$ is calculated as:

$$\upsilon_{i,j} = \begin{cases} 1 & i = j \text{ and } j = 1,2,\cdots n \\ \zeta_j & i = n \text{ and } j = 1,2,\cdots n-1, \\ 0 & otherwise \end{cases} \quad (4)$$

Where the shape of $\Psi^{-1}_{n\times n}$ is similar to matrix $\Psi_{n\times n}$

$$\Psi^{-1}_{n\times n} = \begin{bmatrix} 1 & 0 & \cdots & 0 & 0 \\ 0 & 1 & \ddots & \vdots & 0 \\ \vdots & \ddots & \ddots & 0 & \vdots \\ 0 & \cdots & 0 & 1 & 0 \\ \zeta_1 & \zeta_2 & \cdots & \zeta_{n-1} & 1 \end{bmatrix}_{n\times n} \quad (5)$$

And finally, consider the lower triangular matrix $\Re_{n\times n} = \left[ \beta_{i,j}; i,j = 1,2,....,n \right]$ of the form

$$\beta_{i,j} = \begin{cases} a_j \lambda_j + c_{j-1} \lambda_{j-1} & i = j, j = 1, 2, \cdots n-1 \\ c_j \lambda_j & i = j+1, j = 1, \cdots n-2 \\ d_2 & i = 1, 2, \cdots n-2, j = n \\ d_2 + c_i \lambda_i & i = n-1, j = n \\ \mu_i & i = n, j = n \\ 0 & otherwise \end{cases} \qquad (6)$$

Which could be written as

$$\Re_{n \times n} = \begin{bmatrix} a_1\lambda_1 + c_0\lambda_0 & 0 & 0 & \cdots & 0 & 0 \\ c_1\lambda_1 & a_2\lambda_2 + c_1\lambda_1 & 0 & \cdots & \cdots & \vdots \\ 0 & c_2\lambda_2 & a_3\lambda_3 + c_2\lambda_2 & \ddots & \vdots & \vdots \\ \vdots & \ddots & \ddots & \ddots & 0 & \vdots \\ 0 & \cdots & 0 & c_{n-2}\lambda_{n-2} & a_{n-1}\lambda_{n-1} + c_{n-2}\lambda_{n-2} & 0 \\ d_2 & d_2 & \cdots & d_2 & d_2 + c_{n-1}\lambda_{n-1} & \mu_n \end{bmatrix}_{n \times n} \qquad (7)$$

where the element $\mu_n$ is

$$\mu_n = a_n \lambda_n + (1 + \zeta_{n-1}) c_{n-1} \lambda_{n-1} + d_2 (1 + \sum_{i=1}^{n-1} \zeta_i) \qquad (8)$$

The determinant $\det(\Re)$ of this lower triangular matrix is the product of its diagonal elements:

$$\det(\Re) = \mu_n \prod_{i=1}^{n-1} (a_i \lambda_i + c_{i-1} \lambda_{i-1}) \qquad (9)$$

As the matrix $\Re_{n \times n}$ is a lower triangular matrix, using Eq. (10), its inverse $(\Re)^{-1}_{n \times n} = \left[ \gamma_{i,j}; i, j = 1, 2, \ldots, n \right]$ is obtained in which

$$\gamma_{i,j} = \begin{cases} \dfrac{(-1)^{i+j}\prod_{t=j}^{i-1}c_t\lambda_t}{\prod_{t=j}^{i}(a_t\lambda_t+c_{t-1}\lambda_{t-1})} & i \geq j \text{ and } i = 1,2,\cdots,n-1, j = 1,2,\cdots r \\ \dfrac{\sum_{m=j}^{n-1}\left[(-1)^{m+j+1}\prod_{t=m+1}^{n-1}(a_t\lambda_t+c_{t-1}\lambda_{t-1})\prod_{h=j}^{m-1}c_h\lambda_h\prod_{l=2(n-1)-m}^{n-1}(d_2+c_l\lambda_l)\right]}{\prod_{r=j}^{n-1}(a_t\lambda_t+c_{t-1}\lambda_{t-1})} & i = n \text{ and } j = 1,2,\cdots n-1 \\ \dfrac{1}{\mu_n} & i = n \text{ and } j = n \\ 0 & \text{otherwise} \end{cases} \quad (10)$$

If $\mu_n = 0$, the matrix $A$ is singular, so the matrix has not an inverse matrix which could be found analytically.

### 3.1. Inverse matrix, determinant and solving linear systems of equations

The determinant $\det(A)$ of the matrix $A$ is the product of the determinants of the factor matrices (the determinant of matrix $\Psi$ is one), so

$$\det(A) = \det(\Theta)\det(\Re) \quad (11)$$

where $\det(\Theta)$ and $\det(\Re)$ are formulated in Eq. (16) and Eq. (27). Eq. (29) could be simplified as

$$\det(A) = (-1)^{n-1}\mu_n\prod_{i=1}^{n-1}b_i, \quad (12)$$

Implementing the factorization from Eq. (12), the inverse of matrix $A$ is

$$A^{-1} = \left[(\Re)^{-1}\right]^T \Psi^{-1}\Theta^{-1} \quad (13)$$

Where the elements of $(\Re)^{-1}$ are formulated in Eq. (28), $\Psi^{-1}$ in Eq. (22) and $\Theta^{-1}$ in Eq. (18). However, based on Eq. (31), the linear system $AX = b$ (where $b$ is a vector array) may be solved as $X = A^{-1}b$. As a different approach to solve $AX = b$, given $a_i \lambda_i + c_{i-1} \lambda_{i-1} \neq 0$, $b_i \neq 0$ ($i = 1,...,n-1$) and that the matrix $A$ be nonsingular, we have the following relation based on Eq. (12):

$$AX = b \rightarrow \Psi^{-1} \Theta^{-1} AX = \bar{b} \rightarrow (\Re)^T X = \bar{b} \tag{14}$$

where $\bar{b} = \Psi^{-1} \Theta^{-1} b$. As the matrix $(\Re)^T$ is an upper triangular matrix, using Eq. (32) and the back-substitution procedure yields a solution to the set of equations.

## 5. Discussion

**Theorem**. The matrix $A$ introduced in Eq. (1) could be written as the product of matrices $\Theta$, $\Psi$ and $\Re$ as shown in Eq. (14). This equation represents a decomposition of matrix $A$ into three sparse matrices.

*Proof.*

Considering $a_i \lambda_i + c_{i-1} \lambda_{i-1} \neq 0$, $b_i \neq 0$ ($i = 1,...,n-1$) and that the matrix $A$ be nonsingular the following relation holds

$$A = \Theta \Psi (\Re)^T \leftrightarrow \Psi^{-1} \Theta^{-1} A = (\Re)^T \tag{15}$$

The idea is to prove the right-hand side of the Eq. (33). More specifically, making matrices $\Psi^{-1}$ and $\Theta^{-1}$ in a way that the product of $\Psi^{-1} \Theta^{-1} A$ be $(\Re)^T$.

Now, let matrices $(\Theta_z)_{n \times n} = \left[ (\theta_z)_{i,j} ; i,j = 1,2,....,n \right]$ ($z = 1,2,....,n-1$) be of the form

$$(\theta_z)_{i,j} = \begin{cases} 1 & i = j \text{ and } i = 1,2,\cdots,n \\ \lambda_{z+1} & i = j \text{ and } j = z+1 \\ 1 & i = j+1 \text{ and } j = z \\ 0 & \text{otherwise} \end{cases} \tag{16}$$

where $\lambda_{z+1}$ are presented in Eq. (16) and these matrices are of the form

$$(\Theta_z)_{n \times n} = \begin{bmatrix} 1 & 0 & & & & \cdots & & & 0 \\ 0 & 1 & & & \ddots & \vdots & \cdot^{\cdot^{\cdot}} & & \\ & \ddots & \ddots & & \cdots & 0 & \cdots & & \\ & & 0 & 1 & \ddots & \vdots & & \ddots & \\ \vdots & \cdots & 0 & 1 & \lambda_{z+1} & & & & \vdots \\ & \ddots & \cdots & 0 & 1 & & & & \\ & & & & & \ddots & \ddots & & 0 \\ 0 & & \cdots & & & & 0 & 1 & \end{bmatrix}_{n \times n} \quad (17)$$

Matrices $(\Theta_z)_{n \times n}$ $(z = 1, 2, ...., n-1)$ are considered in a way that multiplying each of them from the left side with matrix $A$ successively removes one of the subdiagonal elements $(b_i)$ and successive multiplications do not undo previously set zeros. The product of multiplying matrices $(\Theta_z)_{n \times n}$ against each other from the left side in the same order as present is matrix $\Theta^{-1}$ (considering $\lambda_1 = 1$)

$$\Theta^{-1} = \Theta_{n-1} \Theta_{n-2} \Theta_{n-3} ... \Theta_z ... \Theta_1 \quad (18)$$

Multiplying matrix $\Theta^{-1}$ from the left side with matrix $A$ consists of a sequence of transformations where each transformation removes one of the subdiagonal elements of matrix Eq. (1).

In order to transform the resultant matrix $(\Theta^{-1} A)$ to the desired upper triangular form $(\Re)^T$, there should be another transformation to remove the element occupying the lower left position of the matrix. This transformation is provided by the matrix $\Psi^{-1}$ which targets this element. If matrix $A$ is an upper triangular matrix $(d_1 = 0)$, matrices $\Psi$ and $\Psi^{-1}$ become the identity matrix (in Eq. (21) all $\zeta_j$ become zero) which agrees with the idea of transforming the matrix to an upper triangular form. Therefore, if the matrix $A$ is a tridiagonal matrix or upper triangular $(d_1 = 0)$, there is no need to calculate $\Psi$ and $\Psi^{-1}$ and the method will be even simpler.

Finally, the upper triangular matrix $(\Re)^T$ is the resultant matrix of multiplying matrices $\Theta^{-1}$, $\Psi^{-1}$ and matrix $A$ subsequently ($\Psi^{-1}\Theta^{-1}A = (\Re)^T$) and based on Eq. (33), Eq. (12) holds.

## 4.1 Non-square matrices and pseudoinverse problem

The formulations of the previous sections are presented for squared matrices for simplicity. However, in most applications, like neural network, the matrix of interest is a non-square matrix. The present algorithm could be implanted to non-squared matrices with the same formulation as:

$$A_{m\times n} = \Theta_{m\times m}\, \Psi_{m\times n}(\Re_{n\times n})^T \tag{37}$$

where $\Theta_{m\times m}$ and $\Re_{n\times n}$ could be calculated with the same formulations mentioned above considering indices $m$ and $n$ and equivalent entries. Matrix $\Psi_{m\times n}$ will have the same size of $A_{m\times n}$ and the same shape as $\Psi$ in the previous sections, but the non-zero row will be at the same spot where there is an entry:

$$\Psi_{n\times n} = \begin{bmatrix} 1 & 0 & \cdots & 0 \\ 0 & 1 & \ddots & \vdots \\ \vdots & \ddots & \ddots & 0 \\ -\zeta_1 & \cdots & -\zeta_{n-1} & 1. \\ 0 & 0 & 0 & 0 \\ \vdots & \vdots & \cdots & \vdots \\ 0 & 0 & \cdots & 0 \end{bmatrix}_{m\times n} \tag{38}$$

And the inverse will be:

$$\Psi^{-1}{}_{n\times n} = \begin{bmatrix} 1 & 0 & \cdots & 0 \\ 0 & 1 & \ddots & \vdots \\ \vdots & \ddots & \ddots & 0 \\ \zeta_1 & \cdots & \zeta_{n-1} & 1. \\ 0 & 0 & 0 & 0 \\ \vdots & \vdots & \cdots & \vdots \\ 0 & 0 & \cdots & 0 \end{bmatrix}_{m\times n} \tag{39}$$

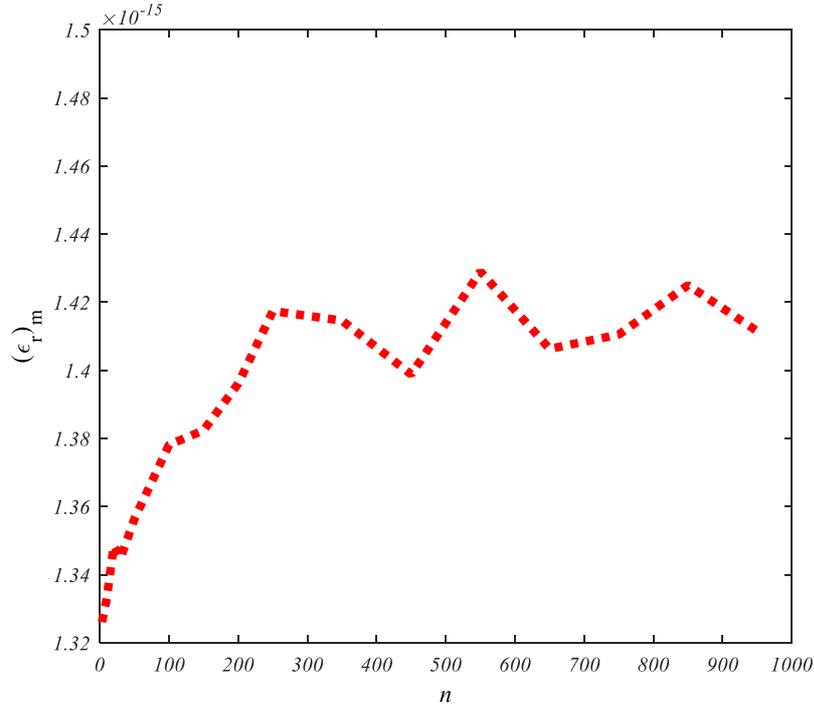

Figure 2. The mean of relative error against the order of matrix

The rest of the algorithm will follow as presented in the previous sections.

**4.2 Validation and efficiency experiment**

We used a computer with Intel (R) Core (TM) i7-2630 QM CPU @ 2.00 GHz, 4.00 GB RAM and MATLAB R21 with Intel(R) Math Kernel Library and IEEE 754-2019 double precision. In this environment, the results of the present work are compared with the results of the well-known LU decomposition matrix inversion method [30]. Inverse of multiple with various sizes ($n$) and random entries are found and the mean of relative errors $(\varepsilon_r)_m$ between two algorithms results are plotted in figure (2). The relative error is measured with infinity-norm of matrices [36] as follows

$$\varepsilon_r = \frac{\left\| A_P^{-1} - A_{LU}^{-1} \right\|_\infty}{\left\| A_{LU}^{-1} \right\|_\infty} \tag{40}$$

where $\varepsilon_r$ is relative error, $A_P^{-1}$ inverse matrix of present work, $A_{LU}^{-1}$ inverse matrix of LU method, and $\left\| \ \right\|_\infty$ denotes infinity-norm operation. As it could be seen from Fig. (1), the results of present work are in excellent agreement with LU algorithm in the range of considered matrix order.

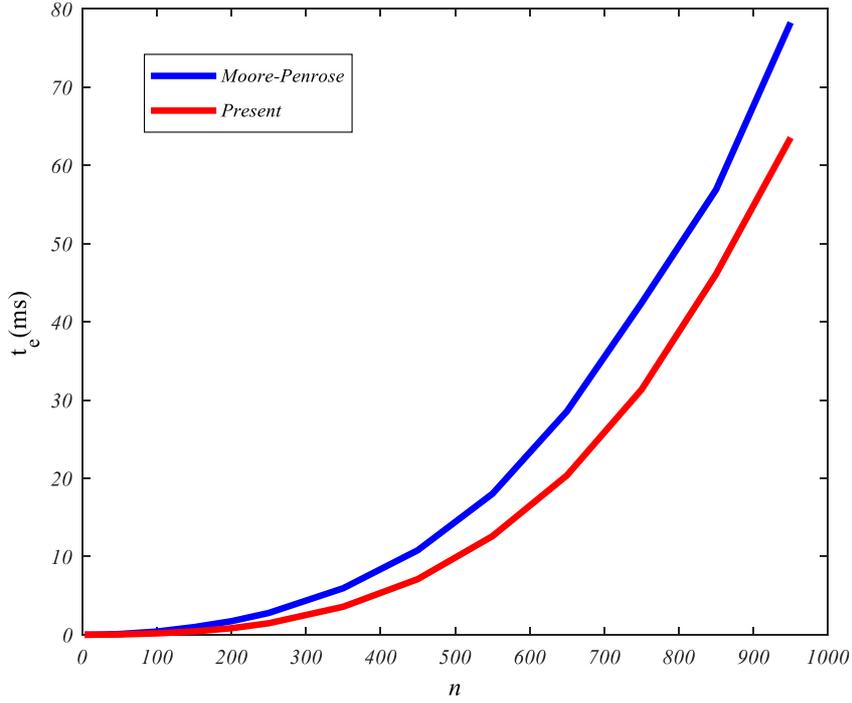

Figure 3. Comparison between the mean of execution time of Moore Penrose and present research methods

Moreover, Calculating the factor matrices aggregately requires $7n-8$ arithmetic operations ($4n-8$ operations for calculating $\Theta$, $2n-3$ for $\Psi$ and $n+3$ operations for $\Re$)). The maximum number of operations for calculating the inverse of factor matrices is $3n^2-2n-12$ arithmetic operations ($3n^2-10n$ for $(\Re)^{-1}$, $2n-3$ for $\Psi^{-1}$ and $6n-9$ operations for calculating $\Theta^{-1}$). The order of required operations for calculating the inverse of the factor matrices is $O(n^2)$. Two matrix multiplications are needed to find the inverse matrix and the best matrix multiplication algorithms require more than $O(n^2)$ operations ($O(n^{2.3728639})$) which means that the order of operations needed for calculating the inverse matrix is dominated by matrix multiplications order (to decrease the number of matrix multiplications, one can combine matrices $\Psi^{-1}$ and $\Theta^{-1}$).

In figure. (3) execution time experiments are done for different random matrices of various orders and elapsed time $t_e$ for calculating inverse matrices using present method is compared with Moore - Penrose inverse method which operates in $O(n^3)$ [30]. In this experiment, all the experiments are carried out on the same hardware, so there is not any hardware-dependent behavior. For each matrix order, random matrices of varying size are generated and the elapsed time for finding their inverse is stored and the mean of output for different matrices of each size are plotted in the figure (3). The results show improvement in timing specially for large $n$. It was observed that the efficiency for lower size matrices are close to each other, but as the size of matrix becomes larger, this our algorithm shows superior performance in matrix calculations.

## 6. Conclusion

This study presented a sparse matrix to be implemented for structuring and optimum calculations of sparse neural networks. The proposed matrix provides enough degrees of freedom for structuring different networks compared to previously implemented methods. Since the proposed algorithm is based on [pseud]inverse training algorithms, efficient methods for calculating inverse, determinant and solving linear systems of equations were formulated. In this methodology, the original matrix ($A$) was factorized into three matrices and to calculate inverse matrices of the factor matrices, another decomposition method is developed and deployed. The closed form [pseudo]inverse of this set of matrices are obtained which subsequently calculates inverse of the matrix $A$. From the general form of the sparse matrix of this study other more specific matrices and neural networks could be driven.